\pdfoutput=1

\documentclass[11pt]{article}

\usepackage[preprint]{acl}

\usepackage{times}
\usepackage{latexsym}

\usepackage[T1]{fontenc}

\usepackage[utf8]{inputenc}

\usepackage{microtype}

\usepackage{inconsolata}

\usepackage{graphicx}

\usepackage{xcolor}

\usepackage{algorithm}
\usepackage{algpseudocode}
\usepackage{amsmath}

\usepackage{listings}
\usepackage{xcolor}

\newcommand{\blue}[1]{{\color{black}#1}}
\title{TaDA: \underline{T}r\underline{a}ining-free recipe for \underline{D}ecoding with \underline{A}daptive KV Cache Compression and Mean-centering}

\author{
 \textbf{Vinay Joshi},
 \textbf{Pratik Prabhanjan Brahma},
 \textbf{Zicheng Liu},
 \textbf{Emad Barsoum}
\\
 AMD
}

\definecolor{codegreen}{rgb}{0,0.6,0}
\definecolor{codegray}{rgb}{0.5,0.5,0.5}
\definecolor{codepurple}{rgb}{0.58,0,0.82}
\definecolor{backcolour}{rgb}{0.95,0.95,0.92}

\lstdefinestyle{mystyle}{
    backgroundcolor=\color{backcolour},   
    commentstyle=\color{codegreen},
    keywordstyle=\color{magenta},
    numberstyle=\tiny\color{codegray},
    stringstyle=\color{codepurple},
    basicstyle=\ttfamily\footnotesize,
    breakatwhitespace=false,         
    breaklines=true,                 
    captionpos=b,                    
    keepspaces=true,                 
    numbers=left,                    
    numbersep=5pt,                  
    showspaces=false,                
    showstringspaces=false,
    showtabs=false,                  
    tabsize=2
}

\begin{document}
\maketitle
\begin{abstract}
The key-value (KV) cache in transformer models is a critical component for efficient decoding or inference, yet its memory demands scale poorly with sequence length, posing a major challenge for scalable deployment of large language models. 
Among several approaches to KV cache compression, quantization of key and value activations has been widely explored. 
\blue{Most KV cache quantization methods still need to manage sparse and noncontiguous outliers separately.}
To address this, we introduce TaDA, a training-free recipe for KV cache compression with quantization precision that adapts to error sensitivity across layers and a mean centering to eliminate separate outlier handling. 
Our approach yields substantial accuracy improvements \blue{for multiple models supporting various context lengths.}
Moreover, our approach does not need to separately manage outlier elements—a persistent hurdle in most traditional quantization methods. 
Experiments on standard benchmarks demonstrate that our technique reduces KV cache memory footprint  to $27\%$ of the original 16-bit baseline while achieving comparable accuracy. Our method paves the way for scalable and high-performance reasoning in language models by potentially \blue{enabling inference for longer context length models, reasoning models, and longer chain of thoughts.}

\end{abstract}
\section{Introduction}


The proliferation of large language models (LLMs) has led to remarkable advancements in natural language processing tasks. However, deploying these models in real-world applications presents significant challenges, particularly concerning memory consumption during inference. A critical component contributing to this issue is the key-value (KV) cache, which stores intermediate representations to expedite autoregressive generation. As sequence length \blue{or number of attention layers increase}, the KV cache's memory footprint expands linearly, often comprising a substantial portion of the total memory usage~\cite{Zhang2023H2O}. The issue is even more pronounced by the advent of large reasoning models and longer inference time thinking where KV cache memory can grow significantly. This poses major challenges on efficient deployment of such LLMs under given hardware constraints.

To mitigate these challenges, early efforts such as multi-query attention (MQA) \cite{shazeer2019fasttransformerdecodingwritehead} and grouped-query attention (GQA) \cite{ainslie2023gqatraininggeneralizedmultiquery} were proposed. MQA reduces the number of key-value heads by sharing a single set of keys and values across all attention heads, thereby decreasing the KV cache size and enhancing inference speed \cite{touvron2023llama2openfoundation}. Despite their benefits, these methods can lead to accuracy degradation and \blue{often require compute intensive full retraining} efforts to recover accuracy \cite{joshi2024qcqaqualitycapacityawaregrouped,yu2024effectivelycompresskvheads}.

\blue{KV cache compression has been approached via different directions, namely 1) token eviction methods that remove non-important tokens \cite{Zhang2023H2O, Liu2023Scissorhands}, 2) quantization of key and value activations \cite{kivi, kang2024gear, hooper2024kvquant}, and 3) low rank approximation of key and value projections matrices \cite{deepseekv2, Chang2024Palu}.}
Prior efforts in KV-cache compression using quantization have laid a robust foundation for reducing memory overhead in LLMs during inference. Early methods in quantization, such as FlexGen \cite{flexgen}, employed 4-bit group-wise quantization to compress both model weights and the KV cache, achieving significant memory savings while maintaining accuracy across diverse tasks. Building on this, KIVI \cite{kivi} introduced a tuning-free 2-bit asymmetric quantization scheme, leveraging per-channel key and per-token value quantization to reduce memory usage. Similarly, GEAR \cite{kang2024gear} combined 4-bit quantization with low-rank and sparse approximations of quantization errors, offering near-lossless performance. QAQ \cite{qaq} proposed quality-adaptive quantization to exploit differing sensitivities in key and value caches, while KVQuant \cite{hooper2024kvquant} pushed boundaries with sub-4-bit quantization, enabling longer context lengths. \blue{Inspired from \cite{kivi}, HuggingFace has enabled 2/4-bit quantization KV cache quantization using Quanto and HQQ libraries \cite{qunato}.}

In this paper, we introduce TaDA, a novel KV cache compression strategy aimed at preserving model accuracy while significantly reducing memory requirements. 
TaDA is motivated by eliminating the need for a separate noncontiguous outlier matrix or low rank and sparse quantization error.
Our approach \blue{simply} mean-centers the key and value activations along the head dimension and quantizes the deviations instead of key and value activations.
During inference, mean-centered activations and quantized deviations are stored instead of \blue{original} key and value activations to reduce KV cache memory overhead.
For attention, computation keys and values are reconstructed from mean-centered activation and quantized deviation.
As we will show empirically, the main motivation behind our approach is that mean-centering reduces the quantization error due to extreme outliers and thus eliminating the need for separate handling of outliers.
TaDA also relies on exploring quantization precision to adapt to error sensitivity across layers via search to further compress KV cache. 
Our method not only alleviates the memory bottleneck but also maintains accuracy levels comparable to the 16-bit \blue{original unquantized} baseline. We explore the efficacy of our approach by evaluating on tasks that necessitate processing longer sequences or more complex structures across different models, demonstrating its versatility and robustness.

\section{Background}
The Transformer architecture, introduced by \cite{vaswani2023attentionneed}, relies on self-attention mechanisms to model relationships between tokens in a sequence. During autoregressive inference, transformers generate tokens sequentially, with each step attending to all previous tokens. To avoid redundant computations, models cache the key and value activations from prior steps, forming the KV cache. While this caching mechanism accelerates inference, it also leads to substantial memory consumption, especially with long input sequences.

To address the memory constraints imposed by the KV cache, researchers have proposed various compression techniques such as multi- or grouped-query attention \cite{shazeer2019fasttransformerdecodingwritehead, ainslie2023gqatraininggeneralizedmultiquery}, dropping of non-important tokens \cite{Liu2023Scissorhands, Zhang2023H2O}, and quantization \cite{flexgen, kivi, kang2024gear, hooper2024kvquant, qaq}. \blue{Among them}, quantization methods reduce the precision of stored keys and values, thereby decreasing memory usage. However, uniform quantization across all heads and tokens can result in information loss and degrade model performance due to extreme and important outliers native to key and value activations. \blue{To the best of our knowledge, unlike for model weights, variable quantization precision across attention layers for KV cache is underexplored.} 

Our proposed method is motivated by outlier-resistant quantization to overcome the need for separate outlier handling. 
By mean-centering the activations along the head dimension and quantize the deviations to low precision, our method demonstrates outlier-agnostic quantization approach for KV cache compression. 
Our method also leverages search to adaptively select quantization precision for different layers based on the error sensitivity. 
TaDA demonstrates substantial reduction in KV cache memory requirements with accuracy comparable to 16-bit original unquantized baseline.

\section{Methodology}
\label{sec:method}
\begin{figure*}
    \centering
    \includegraphics[width=0.95\linewidth]{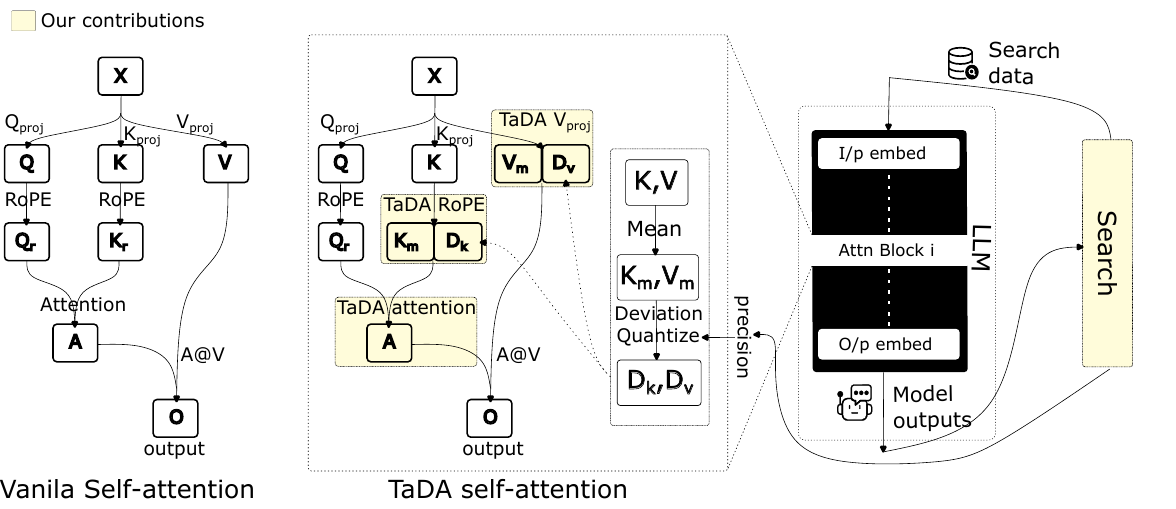}
    \caption{Illustration of TaDA's self-attention mechanism in comparison with vanilla self-attention \cite{vaswani2023attentionneed}. TaDA uses custom Triton kernels to reduce the latency in computing self-attention with compressed forms of key ($K_m$ and $D_k$) and value ($V_m$ and $D_v$) activations (see \ref{sec:method}). Subsequently, flash-decoding kernel is adapted for compatibility with compressed key and value activations in computing self-attention (see \ref{app:triton}). Moreover, TaDA employs random search to adapt quantization precision per layer using a small amount of training set. }
    \label{fig:tada}
\end{figure*}

\blue{In this section we explain our KV cache compression methodology, specifically we maintain a mean-centered key-value activations requiring only $\frac{1}{H}$ ($H$ is the number of attention heads) elements, quantized deviations requiring $(\frac{nbits}{16})^{th}$ the memory and overhead for scaling factors. Mean-centering and deviation computation would be required for each forward pass during inference as shown in Figure \ref{fig:tada}. As an example, for \textit{Llama2-7b} model with 32 heads (each having 128 dimension) and 4-bit quantization precision for deviations, the KV cache memory requirement compared to original unquantized 16-bit baseline is reduced to~$  \frac{1}{32} + \frac{4}{16} + \frac{2}{128} \approx 29\%$.}
\subsection{Mean-centering the key-value activations}
We chose to mean-center the key ($K$) and value ($V$)  activations along the head dimension as follows:

\begin{equation}
    K_{m} = \sum_{i=1:H} K^{i}
\end{equation}

\begin{equation}
    V_{m} = \sum_{i=1:H} V^{i}
\end{equation}

where subscript $m$ stands for mean-centered activation and superscript $i$ denotes the head dimension index. We note that this \blue{is similar in spirit to} what was demonstrated in GQA \cite{ainslie2023gqatraininggeneralizedmultiquery}. However, we 1) do not mean-pool weights but rather activations, and also 2) do not need any further training effort in recovering accuracy. 

\subsection{Computing deviation}
We quantify the deviations for key ($D_{K}^{i}$) and value ($D_{V}^{i}$) activations for an $i^{th}$ head as follows:

\begin{equation}
    D_{K}^{i} = K_{m} - K^{i}
\end{equation}

\begin{equation}
    D_{V}^{i} = V_{m} - V^{i}
\end{equation}

To reduce the memory overhead for storing deviations, we quantize it to lower precision and store it in memory for autoregressive generation. To reduce the overhead of online quantization, we developed Triton kernels to fuse the mean-centering and quantization of deviations in rotary embedding computation for K and projection computation for V  (see appendix \ref{app:triton}). 

\subsection{LLM decoding}
\label{sec:search}
Mean-centered key and value activations and quantized deviation are used to compute attention scores $A$ and output $O$ as follows:

\begin{equation}
    A^{i} = softmax \bigg( \frac{Q^{i} \times (\hat{K}^{i})^{T} }{\sqrt{n}} \bigg)
\end{equation}

\begin{equation}
    O^{i} = A^{i} \times \hat{V}^{i}
\end{equation}

And reconstructed key $\hat{K}$ and value $\hat{V}$ activations are computed as follows:
\begin{equation}
    \hat{K}^{i} = K_{m} - quantize( D_{K}^{i} )
\end{equation}

\begin{equation}
    \hat{V}^{i} = V_{m} - quantize( D_{V}^{i} )
\end{equation}

We developed another Triton kernel to fuse the reconstruction of key and value activations in the flash-decoding kernel (see appendix \ref{app:triton}). This enables TaDA to reduce the overhead of online dequantization unlike in Quanto \cite{qunato}. 
In our experiments, we observed that for \blue{compressing KV cache budget to $\sim 27\%$ or less} suffers from accuracy loss due to insufficient precision for deviations. We employ the following two tailored methods to ensure that we achieve baseline comparable accuracy across different benchmarks and models. 

\textbf{Residual tokens}: We keep track of few past tokens (residual tokens) in high precision without compression. Once the number of past tokens exceeds a certain threshold ($R$), they are compressed and a new set of future tokens are uncompressed and buffered.

\begin{equation}
\label{eq:k_rec}
    \hat{K}^{i}_{r} = cat(  K^{i}[r:], \hat{K}^{i}[:r]  )
\end{equation}

\begin{equation}
\label{eq:v_rec}
    \hat{V}^{i}_{r} = cat(  V^{i}[r:], \hat{V}^{i}[:r]  )
\end{equation}

\blue{The buffer of recent uncompressed ($r~\in~[0,R]$) tokens ($\hat{K}^{i}_{r}$ and $\hat{V}^{i}_{r}$) is concatenated with all previous compressed tokens ($\hat{K}^{i}[:r]$ and $\hat{V}^{i}[:r]$) to obtain key and value tokens for attention computation. This form of retaining uncompressed residual tokens bears resemblance to an implementation demonstrated in \cite{kivi}.}

\textbf{Searching for quantization precision}: We take inspiration from the study \cite{zhang2020acceleratingtrainingtransformerbasedlanguage} that error sensitivity varies across different layers in an LLM. As a result, LLM accuracy is less sensitive to compression in some layers than others. We employ random search by using a small portion of selected samples from a training dataset that is different from the evaluation benchmarks (ensuring there is no data leakage) to identify the optimal sensitivity pattern. This allows us to have variable quantization precision for deviations across different layers and better compress the overall KV cache.

\subsection{Implementation}
To implement TaDA, we have developed three Triton kernels with a goal to minimize the overhead of online mean-centering, quantization, and reconstruction. 
Algorithm \ref{algo:tada} illustrates the steps involved in attention computation using TaDA. 
TaDA shares the same query and key activation computation and applying rotary position embedding (RoPE) \cite{su2021roformer} with original attention implementation \cite{vaswani2023attentionneed}. 
In step 5 of algorithm \ref{algo:tada}, we fuse the RoPE and compression of key activation by developing a custom Triton kernel $CompressV$.
Step 3 demonstrates that instead of computing value activations, we fuse the projection computation with compression for value activations.
Since original flash-attention \cite{flash-attn} is not compatible with TaDA's compressed keys and values, we leverage the flash-decoding kernel from lightllm \cite{lightllm} to create a customized ($TaDAFlashAttn$).
\begin{algorithm}
\caption{Attention computation in TaDA}
\begin{algorithmic}[1]
\Require Input sequence: $X$, Query projection: $W_Q$, Key projection: $W_K$, Value projection: $W_V$
\State $Q$ = Linear($X$, $W_Q$)
\State $K$ = Linear($X$, $W_K$)
\State $V_m, D_V, S_V, M_V = CompressV(X, W_V)$
\State $Q_r$ = RoPE($Q$)
\State $K_m, D_K, S_K, M_K = RoPECompress(K)$
\State $K_s = (K_m, D_K, S_K, M_K)$
\State $V_s = (V_m, D_V, S_V, M_V)$
\State $K_s, V_s = KVCache.update(K_s, V_s)$
\State $O = TaDAFlashAttn(Q_r, Ks, Vs)$
\end{algorithmic}
\label{algo:tada}
\end{algorithm}

\section{Results}
We provide extensive evaluation of our approach and its comparison with recent approaches such as KIVI \cite{kivi} and GEAR \cite{kang2024gear}. The baseline in our results is the uncompressed 16-bit \blue{(BF16 in tables \ref{tab:TaDA-cot} and \ref{tab:TaDA-longb})} KV cache implementation that is, by default, used in all deep learning frameworks.

\subsection{Experimental details}
We evaluate TaDA on various datasets that require longer context for accurate evaluations. We use Llama2-7B \cite{touvron2023llama2openfoundation}, Llama3-8B-it \cite{llama3}, Mistral-7B \cite{mistral7b}, and Mistral-7B-it \cite{mistral7b} models in our evaluations. For layerwise deviation quantization precision search we use a random sample from the training set of hotpotqa dataset on longbench tasks \cite{yang2018hotpotqa}, GSM8k \cite{gsm8k} we used the training set GSM8k. \blue{The use of training set is motivated to simulate true production deployment settings and avoid potential data leakage.} We perform all our evaluations on AMD Instinct$^{TM}$ \blue{MI300} GPUs and each run requires only one GPU. In our Longbench-E evaluations, we used fixed residual length $R$ of 128 tokens and quantization precision for each layer as found to be optimal during the search process. \blue{The search space for quantization precision consists of \{2, 4, 8\}-bits.} For GSM8k experiments, we fixed $R$ to be 32 though.

\subsection{Longbench evaluations}
\begin{table*}[!h]
    \centering
    \begin{tabular}{cccccccc}
\textbf{Model} & \textbf{Method} & \textbf{KV cache} & \textbf{triviaqa} & \textbf{qasper} & \textbf{repobench-p} & \textbf{qmsum} & \textbf{Average} \\
\hline
\hline
Llama2-7b-4k & BF16 & 1.00 & 83.67 & 21.92 & 51.94 & 20.87 & 46.03 \\
\hline
Llama2-7b-4k & KIVI-2-bits & 0.25 & 81.68 & 14.20 & 50.10 & 18.28 & 43.09 \\
\hline
Llama2-7b-4k & KIVI-4-bits & 0.37 & 83.51 & 15.03 & 52.08 & 20.03 & 44.48 \\
\hline
Llama2-7b-4k & GEAR & 0.31 & 84.01 & 15.08 & 52.83 & 20.84 & 45.38 \\
\hline
Llama2-7b-4k & Quanto-2-bit & 0.25 & 81.45 & 12.57 & 43.85 & 19.87 & 41.54 \\
\hline
Llama2-7b-4k & Quanto-4-bit & 0.37 & 83.71 & 22.09 & 51.25 & 21.16 & \textbf{46.11} \\
\hline
Llama2-7b-4k & TaDA & 0.27 & 83.61 & 20.91 & 51.96 & 20.83 & \textbf{45.87} \\
\hline
\hline
Llama3-8b-it-8k & BF16 & 1.00 & 90.21 & 31.20 & 51.19 & 23.52 & 49.51 \\
\hline
Llama3-8b-it-8k & KIVI-2-bits$^*$ & 0.25 & 90.54 & 43.17 & 46.65 & 22.07 & 44.37 \\
\hline
Llama3-8b-it-8k & KIVI-4-bits$^*$ & 0.37 & 90.33 & 44.83 & 52.03 & 22.44 & 45.31 \\
\hline
Llama3-8b-it-8k & Quanto-2-bit & 0.25 & 89.03 & 13.50 & 41.83 & 21.16 & 43.44 \\
\hline
Llama3-8b-it-8k & Quanto-4-bit & 0.37 & 90.89 & 30.19 & 51.08 & 23.06 & \textbf{49.61} \\
\hline
Llama3-8b-it-8k & TaDA & 0.35 & 90.17 & 31.01 & 51.13 & 23.39 & \textbf{49.43} \\
\hline
\hline
Mistral-7b-it-32k & BF16 & 1.00 & 86.29 & 32.57 & 54.08 & 24.22 & 49.27 \\
\hline
Mistral-7b-it-32k & KIVI-2-bits$^*$ & 0.25 & 86.00 & 28.73 & 51.16 & 23.65 & 43.43 \\
\hline
Mistral-7b-it-32k & KIVI-4-bits$^*$ & 0.37 & 86.23 & 29.41 & 51.41 & 24.06 & 43.53 \\
\hline
Mistral-7b-it-32k & Quanto-2-bit & 0.25 & 85.25 & 28.68 & 50.55 & 23.06 & 47.27 \\
\hline
Mistral-7b-it-32k & Quanto-4-bit & 0.37 & 86.23 & 32.09 & 53.87 & 24.64 & \textbf{49.22} \\
\hline
Mistral-7b-it-32k & TaDA & 0.35 & 86.12 & 31.99 & 53.79 & 24.37 & \textbf{49.07} \\
\hline
\hline
Mistral-7b-32k & BF16 & 1.00 & 90.90 & 7.85 & 60.88 & 21.91 & 49.06 \\
\hline
Mistral-7b-32k & KIVI-2-bits$^*$ & 0.25 & 89.63 & 6.92 & 58.99 & 19.71 & 45.85 \\
\hline
Mistral-7b-32k & KIVI-4-bits$^*$ & 0.37 & 89.80 & 7.89 & 58.62 & 20.06 & 46.56 \\
\hline
Mistral-7b-32k & Quanto-2-bit & 0.25 & 90.77 & 5.69 & 54.56 & 21.28 & 45.15 \\
\hline
Mistral-7b-32k & Quanto-4-bit & 0.37 & 90.64 & 7.72 & 60.48 & 21.94 & \textbf{48.85} \\
\hline
Mistral-7b-32k & TaDA & 0.35 & 90.53 & 7.75 & 60.47 & 21.96 & \textbf{48.80} \\
\hline        
    \end{tabular}
    \caption{Evaluation of TaDA's KV cache compression on LongBench eight tasks namely \textit{triviaqa, qasper, trec, samsum, lcc, repobench-p, qmsum}, and \textit{multi-news}. Average is the average across all the eight tasks and only four tasks are shown in the table due to space constraints. $^*$ implies the accuracy numbers are taken from the respective published article. Each model is appended with its context length e.g., Llama3-8b-it-8K model has 8192 context length. We show top-2 performing methods' average accuracy in bold text.}
    \label{tab:TaDA-longb}
\end{table*}

We have evaluated TaDA on the Longbench \cite{bai2024longbench} dataset to study its efficacy on tasks that require a longer context. We report accuracy on the data and KV cache memory requirements normalized to that of 16-bit (BF16) original uncompressed baseline model. We used 1000 random samples from the hotpotqa dataset's training set \cite{yang2018hotpotqa} to \blue{search for an optimal set of precisions per layer}. Table \ref{tab:TaDA-longb} shows evaluation of TaDA, KIVI, and GEAR on multiple Longbench datasets. 
\blue{In general, TaDA achieves the same or better accuracy compared to Quanto, GEAR, and KIVI for lesser cache budget on all the long context tasks with Llama2-7b that is available in MHA configuration.
For pretrained models with GQA (Llama3-8b, Mistral-7b), TaDA performs comparably to Quanto with similar KV cache memory budget. However, unlike Quanto, TaDA offers fused kernel for compression to hide memory transfer latency which can potentially translate into memory and latency savings (see appendix \ref{app:triton}).}

\subsection{Evaluations using chain-of-thought}
\begin{table*}[!h]
    \centering
    \begin{tabular}{cccc}
        \textbf{Model} & \textbf{Method} & \textbf{KV cache} & \textbf{GSM8k}  \\
        \hline
        \hline
Llama2-7b-4K & BF16 & 1.00 & 21.30 \\
\hline
Llama2-7b-4K & KIVI-2-bits & 0.25 & 18.31 \\
\hline
Llama2-7b-4K & KIVI-4-bits & 0.38 & 20.80 \\
\hline
Llama2-7b-4K & GEAR & 0.32 & 21.50 \\
\hline
Llama2-7b-4K & Quanto-2-bit & 0.25 & 13.57 \\
\hline
Llama2-7b-4K & Quanto-4-bit & 0.38 & 20.77 \\
\hline
Llama2-7b-4K & TaDA & 0.27 & \textbf{21.26} \\
\hline
\hline
Llama3-8b-it-8K & BF16 & 1.00 & 67.62 \\
\hline
Llama3-8b-it-8K & GEAR$^*$ & 0.31 & 54.76 \\
\hline
Llama3-8b-it-8K & Quanto-2-bit & 0.25 & 65.65 \\
\hline
Llama3-8b-it-8K & Quanto-4-bit & 0.38 & 42.15 \\
\hline
Llama3-8b-it-8K & TaDA & 0.35 & \textbf{66.73} \\
\hline
\hline
Mistral-7b-it-32K & BF16 & 1.00 & 47.30 \\
\hline
Mistral-7b-it-32K & GEAR$^*$ & 0.31 & 41.93 \\
\hline
Mistral-7b-it-32K & Quanto-2-bit & 0.25 & 36.01 \\
\hline
Mistral-7b-it-32K & Quanto-4-bit & 0.38 & \textbf{45.48} \\
\hline
Mistral-7b-it-32K & TaDA & 0.35 & 44.82 \\
\hline
\hline
Mistral-7b-32K & BF16 & 1.00 & 38.28 \\
\hline
Mistral-7b-32K & KIVI-2-bits$^*$ & 0.25  & 36.01 \\
\hline
Mistral-7b-32K & KIVI-4-bits$^*$ & 0.38  & 37.30 \\
\hline
Mistral-7b-32K & Quanto-2-bit & 0.25 & 26.00 \\
\hline
Mistral-7b-32K & Quanto-4-bit & 0.38 & \textbf{37.83} \\
\hline
Mistral-7b-32K & TaDA & 0.35 & 37.33 \\
\hline

    \end{tabular}
    \caption{Evaluation of TaDA's KV cache compression on tasks requiring chain-of-thought prompting.  $^*$ implies the accuracy numbers are taken from the respective published article. Each model is appended with its context length e.g., Llama3-8b-it-8K model has 8192 context length.}
    \label{tab:TaDA-cot}
\end{table*}

We evaluated TaDA on graduate school math (GSM8k) dataset \cite{gsm8k} to study its efficacy with chain-of-thought (CoT) reasoning, specifically 8-shot CoT, on a mathematical benchmark. As shown in Table \ref{tab:TaDA-cot}, TaDA consistently offers near-baseline (16-bit) accuracy while requiring lower KV cache budget compared to Quanto, KIVI and GEAR for a pretrained model with MHA configuration. With GQA, TaDA's KV cache budget is similar to other methods for better or similar accuracy. 

\subsection{Ablation study}
\begin{figure*}[!h]
    \centering
    \includegraphics[width=0.95\linewidth]{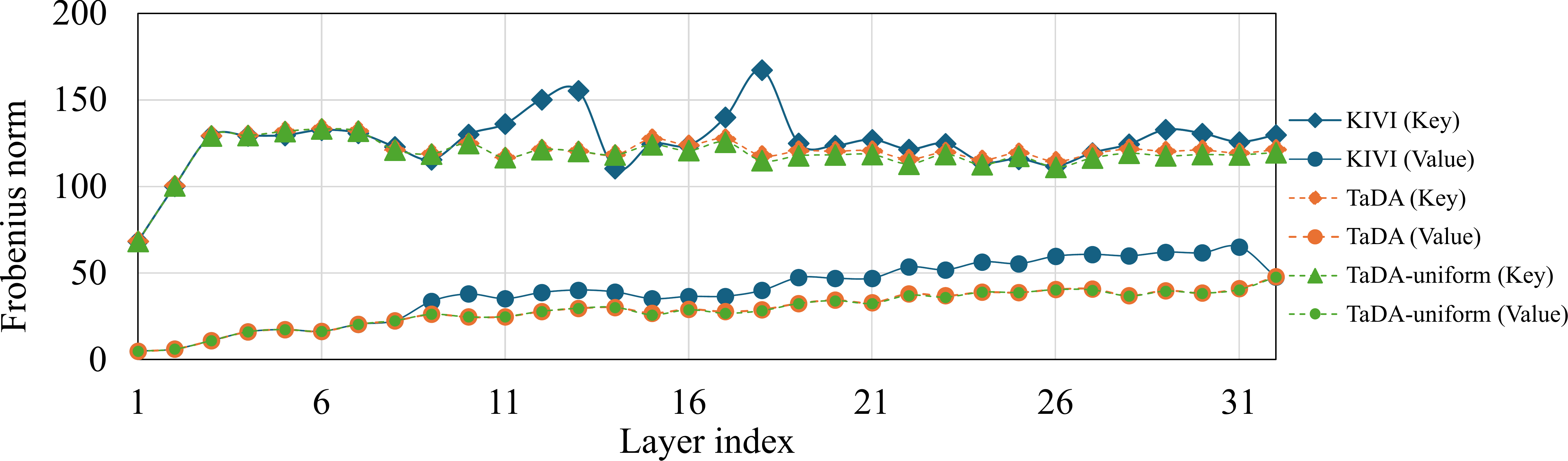}
    \caption{Analysis of key and value activation compression error using Llama2-7B model on hotpotqa dataset's random training set samples. The figure shows Frobenius norm of differences between activations with and without (16-bit uncompressed) compression. TaDA in most layers shows lower Frobenius norm compared to KIVI indicating that TaDA preserves more information compared to KIVI and it is less affected by outliers unlike KIVI. Moreover, label with suffix \textit{uniform} represents TaDA with the same quantization precision across layers. Search does help in reducing the compression error for TaDA but even without search TaDA does better compression than KIVI.}
    \label{fig:frobenius-norm}
\end{figure*}
KIVI and TaDA both approaches do not require separate outlier handling capability unlike other quantization-based KV cache compression methods \blue{(such as \cite{hooper2024kvquant, kang2024gear})} but TaDA consistently outperforms KIVI across different benchmarks and models.
\blue{In our ablation study, we analyze the reconstruction error due to KV cache quantization comparing KIVI and TaDA.}
The reconstruction error is defined as the Frobenius norm of difference between key (and value) activations of quantized and unquantized (baseline) implementations for a subset from the training set of hotpotqa dataset. 
Figure \ref{fig:frobenius-norm} shows the measure of compression error comparing KIVI and TaDA. 
For initial few layers both KIVI and TaDA are comparable but in rest of the layers TaDA has lower Frobenius norm indicating that TaDA's compression preserves more information compared to KIVI. 
The \textit{uniform} suffix in the legend indicates the use of same quantization precision for deviations across layers.
This indicates that quantization precision search largely helps in exploiting layers having lower sensitivity to the error.
As a result, mean-centering and deviation quantization helps in eliminating the need for a separate routine to account for outliers.

\section{Conclusion}
Controlling the KV cache enables online evaluation with extended context lengths, supports bigger model sizes, and allows for larger batch sizes during LLM serving in practical deployments. Our KV cache compression technique TaDA, anchored by mean-centering and deviations stored in adaptively selected low-precision, achieves a synergy along the trade-off between memory efficiency and accuracy that sets it apart from other recent approaches. It achieves near-baseline accuracy with lower KV cache memory budget than other existing quantization methods on long context evaluations. Moreover, our approach sidesteps the complexities of outlier management and delivers a reduction of up to $27\%$ of the baseline memory requirement for KV cache while retaining original accuracy. Ablation studies helped reveal insights into why our approach is more robust to outliers during the quantization process. With custom kernels developed in Triton, TaDA offers an efficient solution for real-world deployment of longer context LLMs and reasoning models.  

\textbf{Limitations and future work}: Our approach relies on using search to find the right quantization precision per layer to achieve appropriate compression. However for each task, we currently make use of a sub-sampled training set that belongs to the same domain but does not contain the same data samples as in the evaluation benchmarks. Such task dependent customization adds some practical challenges for general and scalable deployment. A data-agnostic search or a universal golden dataset for the search would be an interesting solution to this problem but that is left for future exploration.

\bibliography{custom}

\appendix

\section{Efficient Triton kernel for TaDA}
\label{app:triton}
We also developed custom Triton \cite{triton} kernels for TaDA to efficiently realize the gains in KV cache compression. Below, we provide the overview of our kernel design and experiments.

Since autoregressive generation or decoding in LLMs is bottlenecked by memory and especially by KV cache memory transfers at high sequence lengths and bandwidth, Triton kernels enable us to write custom operations to reduce the memory traffic.
A common approach is to fuse multiple operators as is evident from the success of flash-attention \cite{flash-attn}. We fuse the mean-centering and deviation quantization computations with existing operators in the LLM graph. This adds some computational overhead, but removes redundant memory traffic.

\textbf{Compressing key activations}: To hide the latency in computing mean and deviations of key activations, one must fuse these operations with existing ones to eliminate the redundant data transfers from memory.
We achieve this by fusing mean-centering and deviation computation for key activations with rotary position embedding computation. Since RoPE is the most recent computation before updating the KV cache with new tokens, this is the logical operation for fusion. 

\textbf{Compressing value activations}: To hide the latency in computing mean and deviations of value activations, the only obvious operation is linear projection for computing value activation. Unlike key activations, value activations are directly used in attention computation. We fuse the linear projection layer for value activation with mean-centering and deviation quantization to remove redundant memory transfers otherwise.

\textbf{Flash-attention}: Since TaDA stores two components (mean and deviation) per key and value activations, flash-attention kernel cannot be directly used during inference. Flash decoding \cite{hong2024flashdecodingfasterlargelanguage} was proposed as tuned flash-attention kernel specifically for LLM decoding. We adapt the Triton realization of flash decoding from \cite{lightllm} to work with mean-centering and quantized deviation of key and value activations. This helps in removing the overhead of reconstructing key and values by dequantizing them during inference for each input. 

These custom Triton operators enable TaDA to realize its full potential in compressing KV cache and offer better memory consumption and latency for LLM decoding.

\subsection{Performance results}
\begin{table}[!ht]
    \centering
    \begin{tabular}{ccc}
         Method &  Memory (GB) & time/token (ms) \\
         \hline
         \hline
         BF16 & 7.8 & 119.35\\
         \hline
         TaDA (2-bit) & 4.6 & 10.83 \\
        \hline
         TaDA (4-bit) & 6.7 & 40.71 \\
    \end{tabular}
    \caption{Performance measurement of computing single self-attention layer output using TaDA or BF16 with flash-attention-v2 on Llama3.1-70B config ($model\_dim$=8192, $num\_kv\_heads$=8, $num\_attention_heads$=64, $max\_token\_length$=32K).}
    \label{tab:perf}
\end{table}

We measure the execution performance to assess the actual peak memory utilization and latency benefits from executing the TaDA kernel. We run a single self-attention layer using 16-bit original uncompressed (BF16) with flash-attention-v2 \cite{flash-attn} and TaDA for compressing key and value activations.
The dimensions of the self-attention layer match that of Llama3.1-70B \cite{llama3} model, and we run the kernel autoregressively for 32K tokens. The numbers reported are averaged across 100 runs. BF16 in the table refers to baseline PyTorch implementation in brain-float precision format with 16-bits.
Table \ref{tab:perf} shows the peak memory usage and time per token (averaged across 32K tokens and 100 independent runs). TaDA with 2(4)-bit requires only $59\%$ ($85\%$) peak memory compared to BF16. In terms of latency per token, both 2 and 4-bit TaDA require $10\times$ and $3\times$ less compared to BF16.

\section{Quantization precision search}
\begin{table}[]
    \centering
    \begin{tabular}{cccc}
         Model & Accuracy & 4-bit & 2-bit  \\
         \hline
         \hline
         Llama2-7B-4k & 45.90 & 29 & 3 \\
         \hline
         Llama2-7B-4k & 45.87 & 24 & 8 \\
         \hline
         Llama2-7B-4k & 37.31 & 12 & 20 \\
    \end{tabular}
    \caption{Analysis of search candidate outputs on Llama2-7B model for Longbench (hotpotqa's training set). The columns 4-bit and 2-bit indicate the number of layers with that quantization precision for deviations. }
    \label{tab:search}
\end{table}

Our search implementation uses a training set to find optimal candidates for layer-wise quantization precision. We search for \{2, 4, 8\}-bit quantization precision for deviation of both key and value activations. For optimal candidates, we observed that search chooses 4-bit precision for lower layers and 2-bit precision for higher layers. Table \ref{tab:search} shows the analysis of 3 different candidates from search on the Llama2-7B model. As the large number of lower layers use 4-bit precision for deviations, it directly correlates to accuracy improvement.

\end{document}